# A Novel Public Dataset for Strawberry (Fragaria × ananassa) Ripeness Detection and Comparative Evaluation of YOLO-Based Models

Mustafa YURDAKUL[1*], Zeynep Sena BAŞTUĞ[2], Ali Emre GÖK[3], Şakir TAŞDEMİR[4]

[1*] Kırıkkale University, Computer Engineering Dept, Kırıkkale, Türkiye, mustafayurdakul@kku.edu.tr
[2] Kırıkkale University, Computer Engineering Dept, Kırıkkale, Türkiye, 223255002@kku.edu.tr
[3] Hacı Bektaş Veli University, Computer Engineering Dept, Nevsehir, Türkiye, aliemregok@nevsehir.edu.tr
[4] Selçuk University, Computer Engineering Dept, Konya, Türkiye, stasdemir@selcuk.edu.tr

**Abstract:**
The strawberry (Fragaria × ananassa), known worldwide for its economic value and nutritional richness, is a widely cultivated fruit. Determining the correct ripeness level during the harvest period is crucial for both preventing losses for producers and ensuring consumers receive a quality product. However, traditional methods, i.e., visual assessments alone, can be subjective and have a high margin of error. Therefore, computer-assisted systems are needed. However, the scarcity of comprehensive datasets accessible to everyone in the literature makes it difficult to compare studies in this field. In this study, a new and publicly available strawberry ripeness dataset, consisting of 566 images and 1,201 labeled objects, prepared under variable light and environmental conditions in two different greenhouses in Turkey, is presented to the literature. Comparative tests conducted on the data set using YOLOv8, YOLOv9, and YOLO11-based models showed that the highest precision value was 90.94% in the YOLOv9c model, while the highest recall value was 83.74% in the YOLO11s model. In terms of the general performance criterion mAP@50, YOLOv8s was the best performing model with a success rate of 86.09%. The results show that small and medium-sized models work more balanced and efficiently on this type of dataset, while also establishing a fundamental reference point for smart agriculture applications.

## 1. Introduction

Strawberry (Fragaria × ananassa) is a widely consumed fruit with high economic value all over the world. Strawberries, which are commercially cultivated nowadays, first appeared in Europe in the eighteenth century and were derived from the natural hybridization of Fragaria virginiana, a species native to North America, and Fragaria chiloensis, another species native to South America(Hummer & Hancock, 2009; Staudt, 1999). Initially cultivated in France, thanks to its large sizes, aroma characteristics, and potential for high yields, it quickly spread from Europe to the American continent and then to Asia. Over time, numerous commercial strawberry varieties have been cultivated that can adapt to different climates and growing conditions(López-Aranda et al., 2011).

Strawberries are one of the most important foods for a healthy diet as they are full of nutrients. In addition to being a fruit that supports the immune system with its high vitamin C content, phenolic compounds, and antioxidant capacity, it also provides important benefits for digestive health thanks to its dietary fiber content(Alwazeer & Özkan, 2022; Rana et al., 2020). Strawberries are a food that is very attractive for both fresh consumption and processed

Corresponding Author: Mustafa Yurdakul, Kırıkkale University Computer Engineering Departmant, mustafayurdakul@kku.edu.tr

products in the food industry due to their nutritional properties. (Çelik et al., 2026). However, the physical structure and water content of the strawberry also cause significant challenges in post-harvest processes(Dziedzinska et al., 2018; Luksiene & Buchovec, 2019). In particular, its very short shelf life makes it critical to accurately and consistently determine the fruit's ripeness level during harvest and marketing(Gol et al., 2013; Hu et al., 2012).

Harvesting fruit at the right time is of great importance. If fruit is picked before it is fully-ripe, it cannot develop its expected aroma; its color will not be sufficiently vibrant, and its nutritional value will not reach the required level. On the other hand, if harvesting is delayed, the fruit becomes overripe; this increases the risk of spoilage, bruising, and quality loss during transportation and storage. As a result, not only do producers and the supply chain face economic losses, but there are also significant fluctuations in the quality of the product that reaches the consumer. When global strawberry production is examined, countries such as China, the United States, Mexico, Turkey, and Spain emerge as the leading producers(Benlioglu et al., 2014; Mok et al., 2014; Simpson, 2018). With the spread of greenhouse cultivation, large quantities of strawberries are produced year-round in greenhouses, making it even more important to determine ripeness using fast and reliable methods.

In traditional cultivation processes, the ripeness of strawberries is determined by visual observation based on fruit characteristics such as color, brightness, and surface texture. However, the evaluation process is subjective and depends on the experience of the person conducting the assessment. Therefore it does not provide a standard quality benchmark(Morillo et al., 2015; Van Delm et al., 2016; Yang & Kim, 2023). Moreover manually checking thousands of fruits, in large-scale greenhouses, is both time-consuming and increases labor costs. Also prolonged visual assessment increases human-induced errors, while variable lighting conditions and environmental factors can cause differences in perception. It can lead to inconsistencies in product quality and significant economic losses in the marketing process.

In this context, the automatic, rapid, and objective detection of strawberry ripeness has emerged as an important research topic for modern agricultural practices(Yurdakul et al., 2024; Yurdakul, Uyar, & Taşdemir, 2025). In particular, computer vision and DL based approaches overcome the limitations of human observation-based methods, achieving high accuracy (Uyar et al., 2024; Yurdakul, Uyar, et al., 2025). In addition, strawberry ripeness detection stands out as a critical component for robotic applications and autonomous harvesting systems developed within the scope of smart farming systems. The ability of autonomous harvesting robots to harvest the right fruit at the right time is directly dependent on the accuracy of ripeness detection. Therefore, the development of real-time, highly accurate, and computationally efficient automatic ripeness detection systems is of great importance for both academic research and industrial applications.

## 2. Related Works

Advances in computer vision and deep learning in recent years have significantly accelerated research on automated quality assessment and ripeness detection in agricultural production. Strawberry-specific research has been structured around a variety of problem definitions, as follows:

(i) ripeness detection,
(ii) simultaneous fruit–stem detection,
(iii) segmentation-based approaches,
(iv) harvest time prediction,
(v) physical defect and deformation analysis,
(vi) models resilient to variable and challenging environmental conditions.

However, in terms of the model architectures, datasets, and performance evaluation metrics used, there are significant methodological differences in the literature.

Transformer-based architectures are prominent in studies addressing the problem of ripeness detection in field environments. Zhao et al.(Zhao et al., 2025) proposed the transformer-based FruitQuery model to perform in-field ripeness detection for strawberries and peaches. The model achieved an AP value of 67.02% with 14.08 million parameters, outperforming the YOLOv8, YOLOv9, and YOLOv10 models. However, a decrease in accuracy for small and distant objects creates limitations in terms of generalizability under real greenhouse conditions.

Yang et al.(Yang et al., 2025) proposed the PDSE-DETR model to detect ripeness of strawberry in greenhouse environments. PDSE-DETR increased accuracy by 2.1% while reducing parameter and computational costs by 30.2% and 30.7%, respectively. However, inference speed of the proposed model remains low compared to other YOLO-based models. These studies demonstrate that transformer architectures can increase accuracy, but improvements are still needed in terms of the speed-efficiency balance.

Since efficiency is a important in real-time applications, YOLO-based approaches are proposed in the literature. Yu et al.(Yu et al., 2025) proposed the Ripe-Detection model for ripeness detection system under complex environmental conditions. The authors achieved 96.4% mAP50, providing a 3.9%–13% improvement over YOLO models. However, a tendency to produce false positives in complex backgrounds has been reported.

Ma et al.(Ma et al., 2025) proposed the YOLOv11-GSF model to for ripeness detection system; the model achieved a 1.8% increase in AP, a 1.3% increase in accuracy, and a 2.1% increase in recall compared to YOLOv11. However, comprehensive validation under dynamic field conditions has not been performed.

Wang et al. (Wang et al., 2025) presented a YOLOv11-based model to develop a real-time rotten fruit detection and smart sorting system, achieving an mAP of 83.6%; however, the use of RGB images alone limited the detection of early-stage low-contrast defects.

Jiang et al. (Jiang et al., 2025) proposed a YOLOv11-based model and geometric distance metrics to evaluate strawberry deformation and symmetry analysis, achieving 91.11% precision and 92.9% AP50. They reported a need for real-time performance optimization in high-density planting areas.

Segmentation and multi-stage approaches also play an important role in the literature.

Crespo et al. (Crespo et al., 2025)optimized the Mask R-CNN model with TensorRT to provide efficient segmentation suitable for real-time applications, achieving 83.45% mAP and 4 FPS before optimization and 83.17% mAP and 25.46 FPS after optimization.

Lin et al.(Lin et al., 2025) developed a two-stage (U-Net + TBAF) model to predict strawberry harvest dates, achieving 0.977 mIoU in segmentation and 0.859 accuracy and F1-Score in harvest time prediction. Sun et al. (Sun et al., 2026) proposed the SRR-Net model to make ripeness predictions resistant to variable lighting conditions and achieved a 0.037 MAE value, representing an improvement of over 70% compared to Mask R-CNN; however, it showed high dependence on light intensity normalization.

Wu et al.(Wu et al., 2026) developed a semi-supervised SSEFNet model to increase harvest efficiency and achieved 91.1% precision on complex datasets; however, the model's dependence on high-quality labeled data limits its generalization.

Nagaki et al. (Nagaki et al., 2025) proposed an EfficientNetV2-based model to predict ripeness based on ranking; they achieved $R^2 = 0.84$ and overall accuracy of 0.83, but dependence on panelist consistency was noted as a significant limitation.

When the studies summarized in Table 1 are evaluated together, three main limitations emerge:

(1) Most studies focus on a single architecture and do not systematically compare different model generations under the same experimental conditions.
(2) The data sets used are mostly private and inaccessible, which reduces comparability and reproducibility.
(3) Although accuracy metrics are reported, application-critical criteria such as computational cost, number of parameters, and real-time performance are not comprehensively analyzed.

**Table 1.** Summary of recent DL–based studies on strawberry ripeness detection and related tasks

| Study | Objective | Proposed Method | Main Results | Limitations |
|---|---|---|---|---|
| Zhao et al. | In-field ripeness detection of strawberries and peaches | Transformer-based FruitQuery (Instance Segmentation) | 67.02% AP with 14.08M parameters; outperformed YOLOv8–v10 | Reduced accuracy for small-scale or distant instances compared to heavier architectures. |
| Wu et al. | Simultaneous detection of strawberry fruit and stem | Semi-supervised SEFNet (SSEFNet) | 91.1% precision on complex datasets | High dependency on quality labeled data for morphological generalization. |
| Wang et al. | Real-time strawberry bruise detection and intelligent sorting | YOLOv11-based detection model | 83.6% mAP; superior to Mask R-CNN and YOLO variants | RGB imaging limits detection of early-stage, low-contrast bruises. |
| Crespo et al. | Efficient strawberry segmentation for real-time applications | Mask R-CNN with TensorRT optimization | Pre-optimization: 83.45% mAP, 4 FPS; Post-optimization: 83.17% mAP, 25.46 FPS | Segmentation performance degrades under heavy fruit overlap and complex foliage. |
| Lin et al. | Strawberry harvest date prediction | Two-stage fusion model (U-Net + TBAF) | 0.977 mIoU (segmentation); 0.859 accuracy and F1-score | Accuracy sensitive to micro-environmental fluctuations in plant factory settings. |
| Sun et al. | Light-resilient strawberry ripeness estimation | SRR-Net (YOLOv8/YOLOv11-based regression) | MAE = 0.037; >70% improvement over Mask R-CNN | Heavily dependent on light-intensity normalization; reduced robustness in unpredictable weather. |

| Yang et al. | Ripeness detection in greenhouse environments | PDSE-DETR (Improved RT-DETR) | 2.1% accuracy improvement; 30.2% parameter and 30.7% FLOPs reduction | Inference speed trails SOTA YOLO models, constraining ultra-fast applications. |
|---|---|---|---|---|
| Yu et al. | Robust strawberry ripeness detection under challenging conditions | Ripe-Detection | 96.4% mAP50; 3.9–13% improvement over YOLO models | Susceptible to false positives in complex backgrounds; lacks dataset diversity. |
| Ma et al. | Efficient and accurate ripeness detection | YOLOv11-GSF | AP +1.8%, accuracy +1.3%, recall +2.1% vs. YOLOv11 | Lacks validation under dynamic field conditions with severe occlusion. |
| Jiang et al. | Strawberry deformity detection and symmetry analysis | YOLOv11 + geometric feature analysis | 91.11% precision; 92.9% AP50 | Limited validation in high-density planting; real-time latency requires optimization. |
| Nagaki et al. | Ripeness ranking prediction | EfficientNetV2-based ranking model | $R^2 = 0.84$; overall accuracy = 0.83 | Dependent on panelist consistency; lacks cross-cultural sensory validation. |

In this study, the gap in the literature is addressed in two ways. First, a publicly available dataset collected from two different greenhouses and under varying light conditions is presented. Second, the YOLOv8, YOLOv9, and YOLO11 architectures are evaluated under the same dataset and experimental conditions; accuracy metrics (Precision, Recall, mAP@50) along with computational cost (FLOPs, number of parameters, FPS) are analyzed holistically. In this way, both methodological consistency was ensured and the architecture scale–performance relationship in the context of strawberry ripeness detection was systematically revealed.

## 3. Material and methods

The study has designed an end-to-end DL workflow for strawberry ripeness detection. The process consists of collecting images from two different greenhouses, manual annotation based on bounding boxes, dividing the data into training, validation, and test sets, and training the YOLOv8, YOLOv9, and YOLO11 architectures under the same experimental conditions. The models were evaluated using the metrics of correctness (Precision, Recall, mAP@50) and computational efficiency (FLOPs, number of parameters, FPS), and the most suitable model was determined based on the correctness–cost balance. The experimental flow of the study is shown schematically in Figure 1.

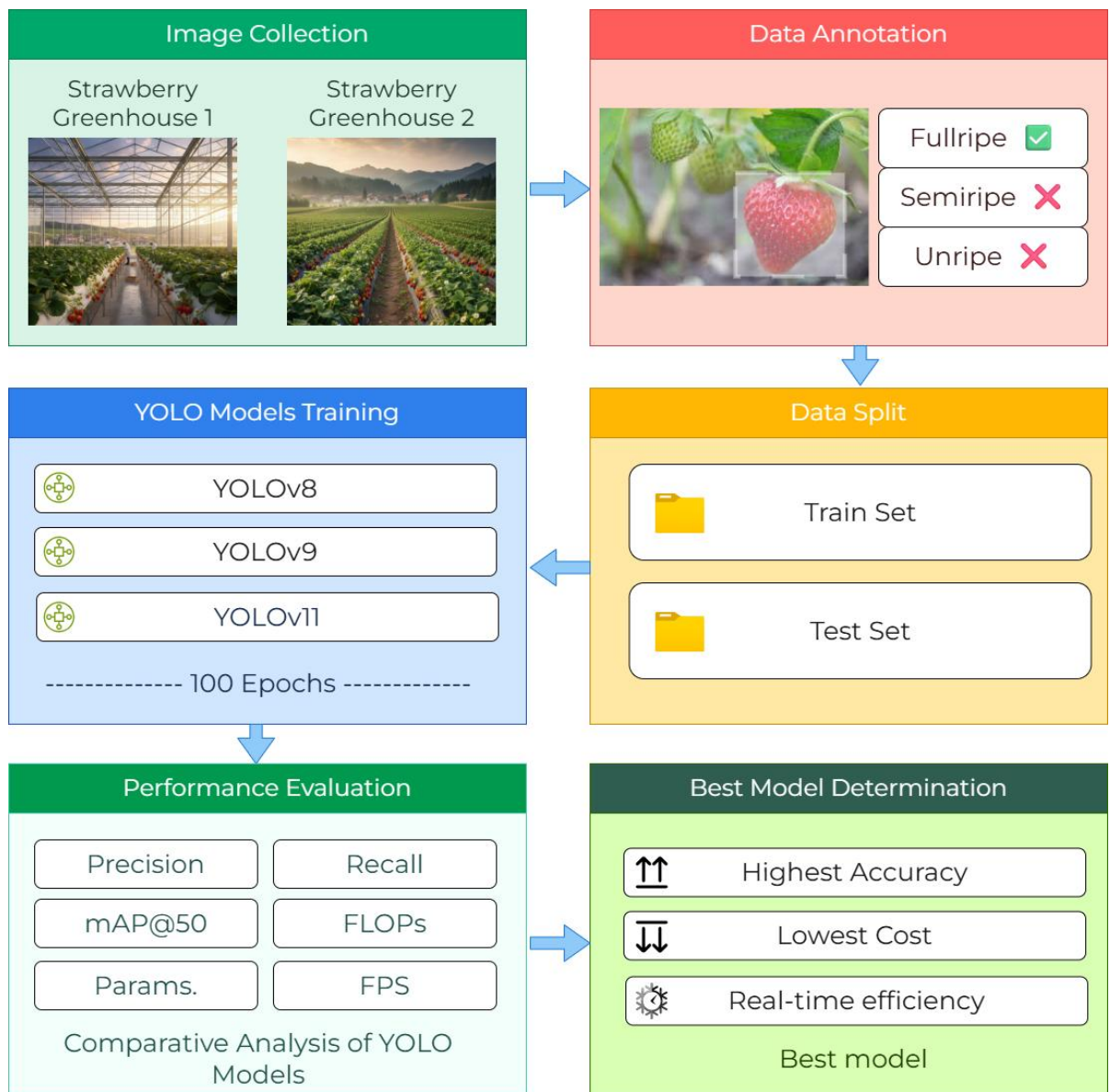

**Figure 1.** Schematic diagram of the proposed experimental workflow

## 3.1. Dataset

The dataset used in this study was created by the authors using a Samsung NX camera and a mobile phone camera to reflect different imaging conditions. The images were collected from two different greenhouse located in the Kayseri, Türkiye. The selection of different greenhouse ensured that factors such as background, density, lighting, and cultivation types were reflected in the dataset.

The images were captured at different times of the day and under varying natural light conditions during the data collection process. Different scenarios, such as direct sunlight, partial shading, and diffuse light inside the greenhouse, were considered.

The created dataset consists of a total of 566 images. The dataset is divided into two subsets: training and testing. There are 465 images in the training set and 101 images in the test set. Distribution information for the subsets is presented in Table 2.

**Table 2.** Distribution of images across training, validation, and test subsets

| Train | Test | Total |
|---|---|---|
| 465 | 101 | 566 |

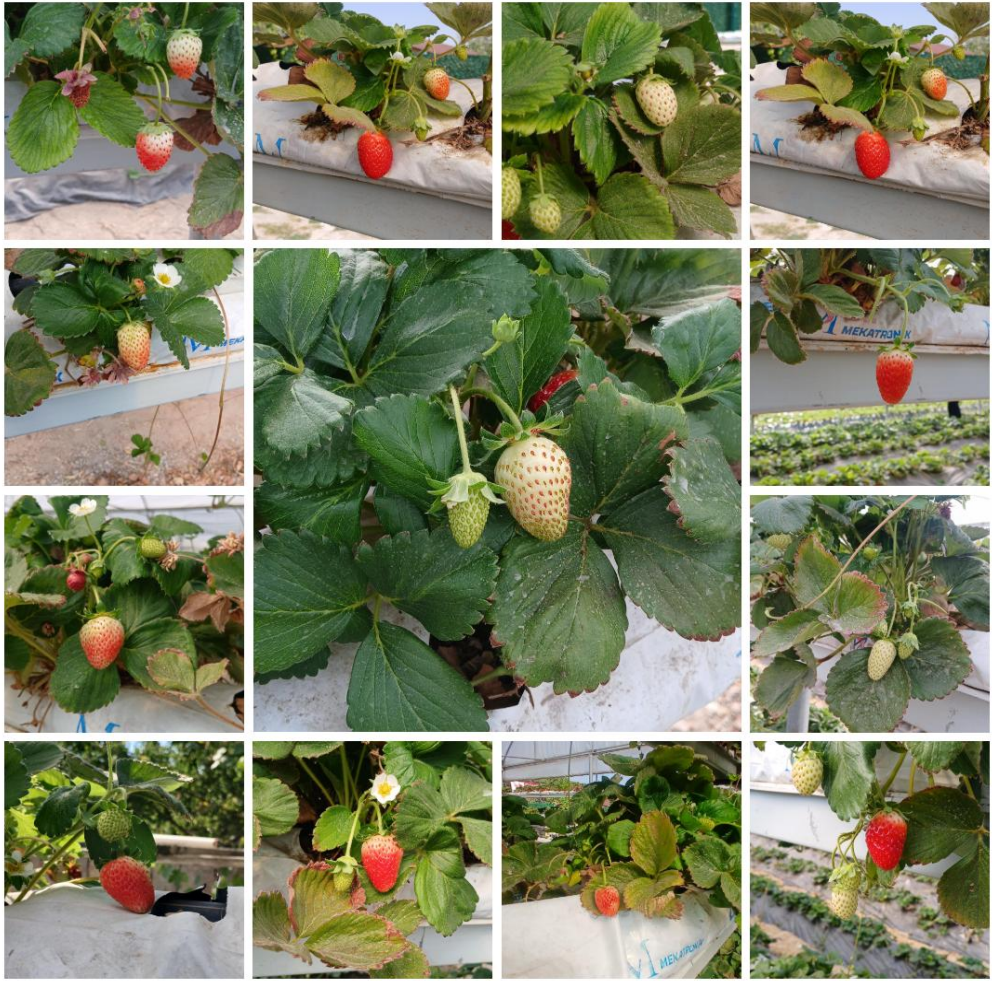

**Figure 2**. Sample images of ripe and unripe strawberries collected from two different greenhouses

The dataset contains a total of 1,734 object annotations. Images were labeled according to three different ripeness classes—fully-ripe, semi-ripe, and unripe. The distribution of images across classes is detailed in Table 3. In total, there are 520 fully-ripe, 247 semi-ripe, and 967 unripe strawberry annotations.

**Table 3.** Number of annotated objects per class (fully-ripe, semi-ripe, unripe) in training, validation, and test sets

|  | **Fully-ripe** | **Semi-ripe** | **Unripe** | **Total** |
|---|---|---|---|---|
| **Count** | 520 | 247 | 967 | 1734 |

All annotation were performed manually using the bounding-box method, and annotation accuracy was carefully checked. Sample images from the dataset are presented in Figure 2. In addition, the dataset created to increase the reproducibility of the study and contribute to the literature has been made publicly available. It can be accessed the from

https://www.kaggle.com/datasets/mahyeks/multi-class-strawberry-ripeness-detection-dataset.

**3.2. YOLO Models**

Object detection algorithms are categorized as one-stage and two-stage methods(Yurdakul & Taşdemir, 2025). In two-stage methods, candidate regions are first generated, and then these regions are classified. Although two stage methods generally provide high accuracy, they are not preferred in real-time applications due to their high computational cost(Yurdakul & Tasdemir, 2025). On the other hand, single-stage methods, predict object locations and class information in a single step. They work faster and can operate with relatively lower hardware requirements (Yurdakul & Tasdemir, 2025). The YOLO architecture has significant potential in agricultural fields such as greenhouse environments, open field applications, and mobile robot systems, thanks to its real-time processing advantage.

The YOLO family has been continuously improved since its first release. In this study, the current versions of YOLOv8, YOLOv9 and YOLO11 are tested comparatively. The basic architectural features of each model are summarized below.

*YOLOv8*

The YOLOv8(Ultralytics, 2023) architecture is constructed on the backbone–neck–head paradigm and is designed to have a lighter, more modular, and efficient structure compared to previous YOLO versions. The architectural diagram of YOLOv8 is shown in Figure 3.

In the backbone stage, the input image is transformed into multi-level feature maps through sequential convolution layers and C2f (two-convolution Cross Stage Partial) blocks. Compared to traditional CSP-based structures, the C2f block provides a more efficient gradient flow, increasing representational power in deep layers while also improving parameter efficiency. The SPPF (Spatial Pyramid Pooling – Fast) module, located in the deeper levels of the backbone, combines pooling operations with different receptive fields to enrich contextual information and increase the network's sensitivity to both small and large-scale objects. It allows global contextual information to be integrated into the model without adding extra computational load.

A structure similar to PANet is adopted in the Neck stage. In this stage, feature maps from the upper and lower levels are combined through Upsample and Concat operations. Thanks to multi-scale feature fusion, objects at different resolutions are better represented. The C2f blocks used within the Neck ensure that the combined features are processed in a discriminative and efficient manner while keeping computational costs low.

In the Head stage, YOLOv8 uses an anchor-free detection paradigm, which is different from previous YOLO versions. Feature maps generated at each scale are passed to separate Detect layers, where class probabilities and bounding box predictions are obtained directly through regression (Ultralytics, 2023).

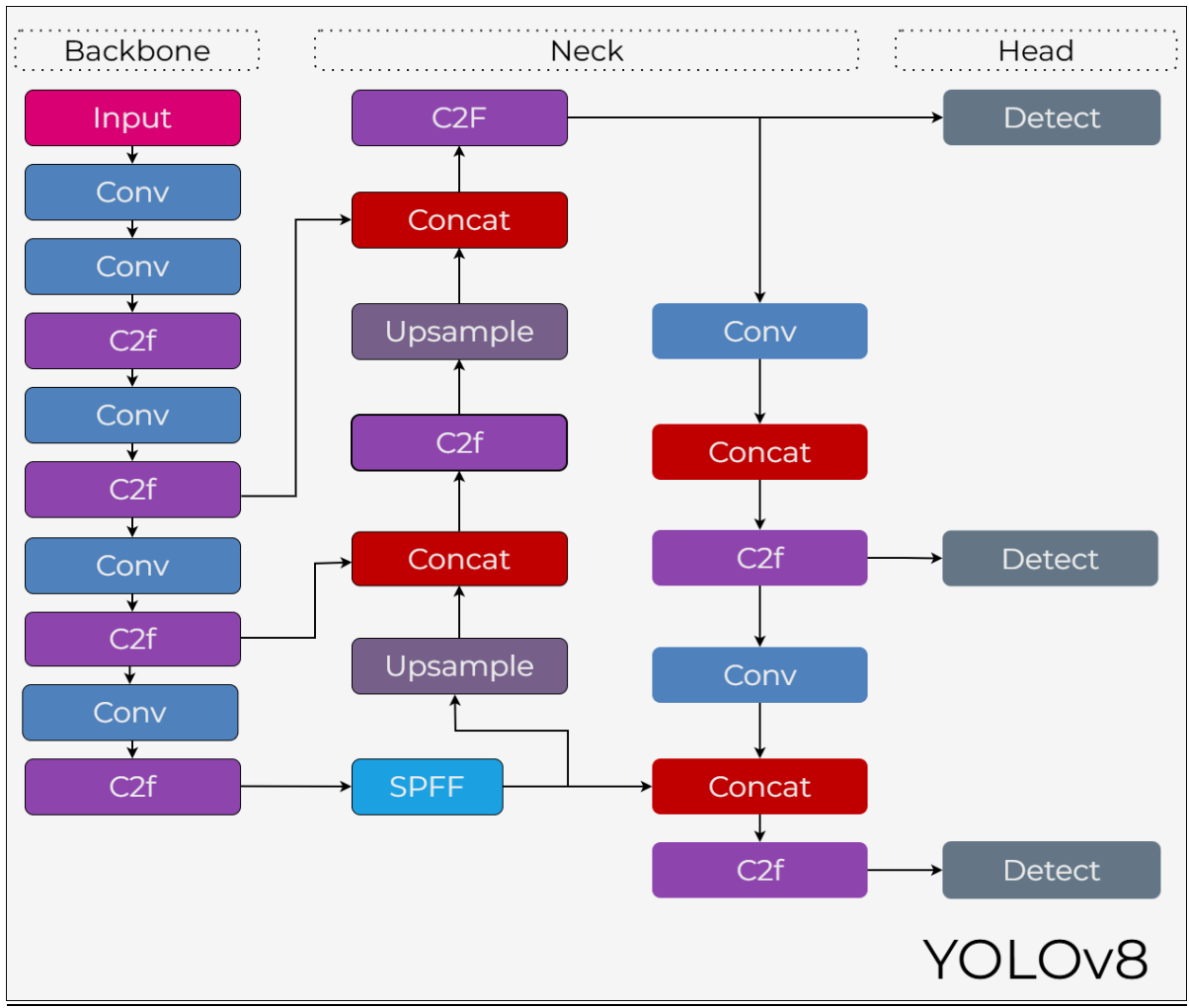

**Figure 3.** Schematic representation of the YOLOv8 architecture

YOLOv9

The YOLOv9(WongKinYiu & Ultralytics, 2024) architecture is based on the backbone–neck–head distinction, as in YOLOv8. However, it has components to reduce information loss in deep layers and improve gradient flow. The architectural structure of YOLOv9 is shown in Figure 4.

In the backbone stage, the input image is transformed into multi-level feature representations through sequential convolutional layers and RepNCSPELAN4 blocks. The RepNCSPELAN4 block, based on the reparameterization principle, offers a multi-branched representation, while transforming into a single-branched and lighter structure. The SPPELAN module, located in the deeper layers of the backbone, improves contextual representation by combining spatial information obtained from different receptive fields.

When transitioning from the backbone to the neck stage, CBLinear layers are used to more efficiently combine features from different hierarchical levels, and channel sizes are balanced. In the neck stage, the CBFuse (Cross-Branch Feature Fusion) module, one of the distinctive components of YOLOv9, plays an important role. The CBFuse structure reduces information loss and increases the effectiveness of multi-scale representations by directly and selectively merging feature maps at different resolutions. At this stage, a PAN-like multi-scale fusion structure is created through Upsample and Concat operations. RepNCSPELAN4 blocks ensure that the fused features are processed in a computationally efficient and discriminative manner. In the Head stage, feature maps obtained from different scales are fed into multiple Detect layers to perform multi-scale object detection(WongKinYiu & Ultralytics, 2024).

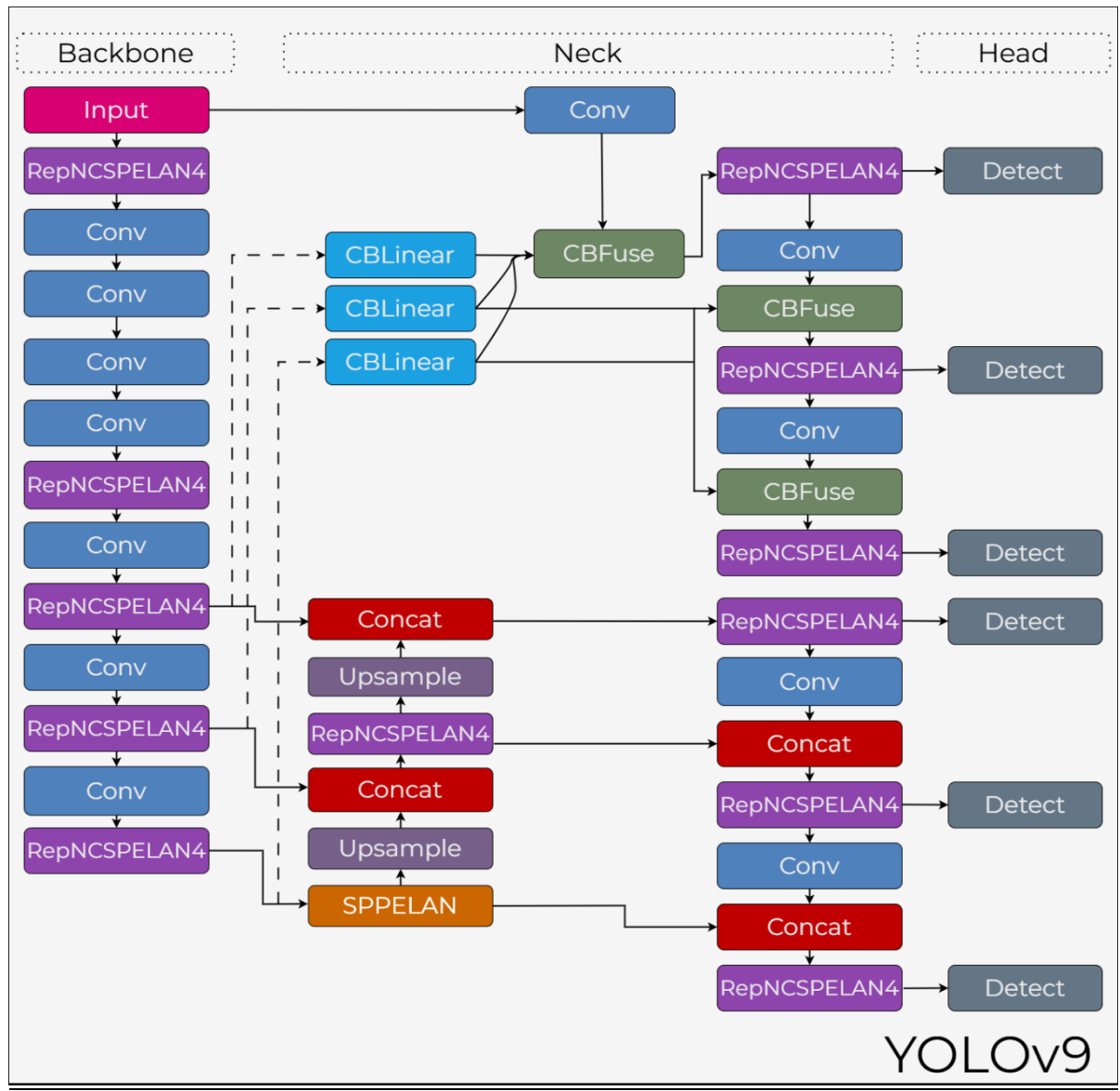

**Figure 4.** Schematic representation of the YOLOv9 architecture

YOLOv11

The YOLOv11 (Ultralytics, 2024) architecture is also relies on the backbone–neck–head structure. A schematic representation of the YOLOv11 architecture is shown in Figure 5.

In the backbone stage, the input image is converted into multi-scale feature maps through sequential convolutional layers and optimized C3k2 blocks. Compared to the C2f structure used in previous versions, the C3k2 block decreases the number of parameters by using smaller convolutional kernels. The Spatial Pyramid Pooling–Fast (SPPF) block, located at the backbone, combines contextual information obtained from different receptive fields to extract global features. The following Convolution Block with Parallel Spatial Attention (C2PSA) module enables the network to focus on critical regions.

In the Neck stage, feature maps from lower and upper levels are combined using Upsample and Concat operations. The C3k2 blocks, which are used repeatedly within the Neck, enable the combined features to be processed with low computational cost and high discriminative power.

In the Head stage, feature maps generated at different scales are fed into separate Detect layers to perform multi-scale object detection.

Ultimately, through the combined use of the C3k2, SPPF, and C2PSA blocks, YOLOv11 presents a scalable and efficient architecture. With these features, it offers a model for real-time applications on both edge devices and high-performance GPU systems(Ultralytics, 2024).

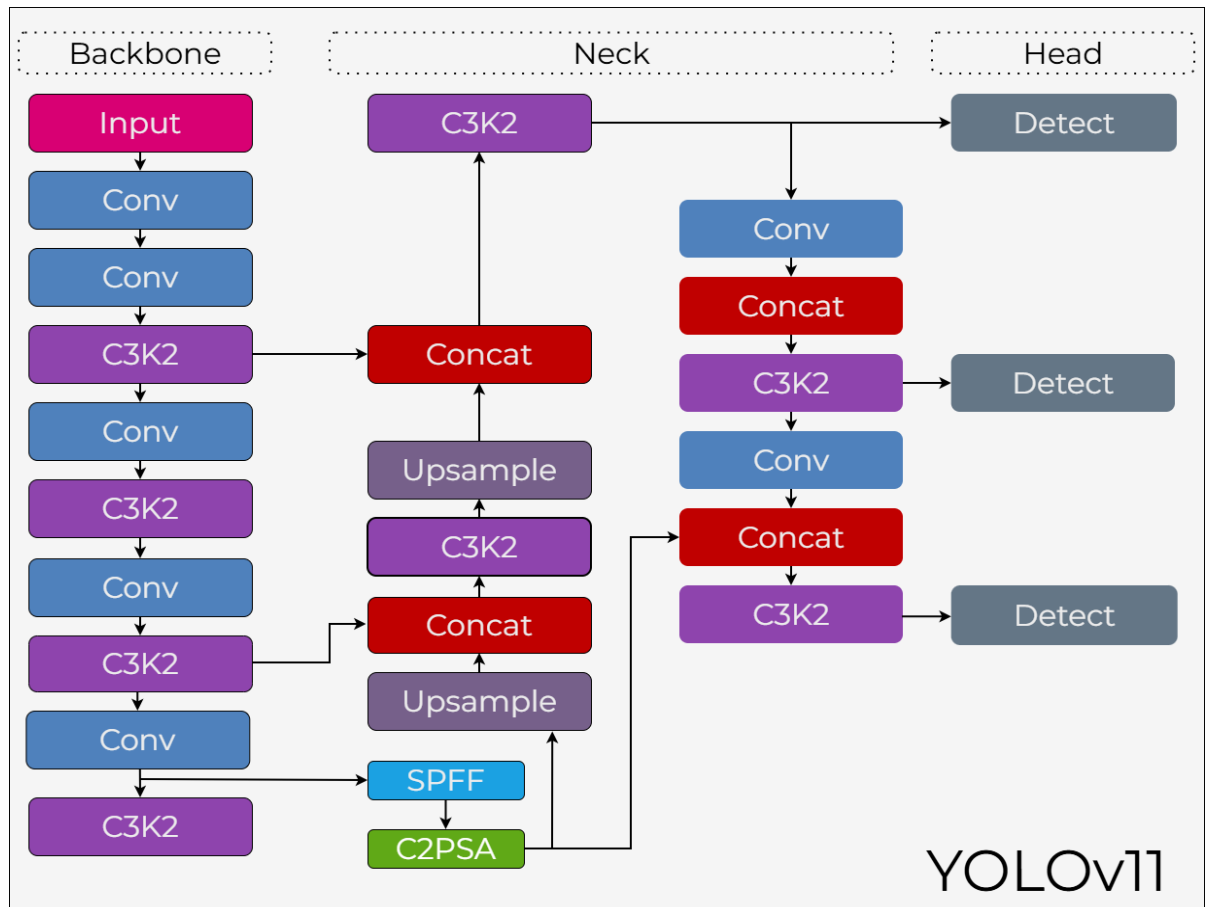

**Figure 5.** Schematic representation of the YOLOv11 architecture

### 3.3. Evaluation metrics

In this study, different metrics were used to evaluate the performance of YOLO-based models in terms of the accuracy and computational efficiency. Precision measures how many of the model's positive predictions are correct. Recall Indicates how many true positive examples are correctly captured by the model.

Mean Average Precision (mAP@50) is the average of the AP values calculated for all classes. In this study, the mAP@50 value was used with an IoU threshold of 0.5. mAP is a widely accepted benchmark in object detection models. Floating Point Operations (FLOPs) indicate the number of mathematical operations the model performs during prediction. Frames Per Second (FPS) indicates how many images the model can process per second. The mathematical equations for the metrics are provided in Equations 1–5.

$$Precision = \frac{TP}{TP + FP} \tag{1}$$

$$Recall = \frac{TP}{TP + FN} \tag{2}$$

$$AP = \int_0^1 P(R)dR \tag{3}$$

$$mAP@50 = \frac{1}{N}\sum_i^N AP_i \tag{4}$$

$$FLOPs = 2 \times (C_{in} \times K^2 \times C_{out} \times H_{out} \times W_{out}) \tag{5}$$

$$FPS = \frac{N}{t_{total}} \tag{6}$$

## 4. Results

A comprehensive experimental analysis was performed on the dataset collected within the scope of this study. Different scale versions of YOLOv8, YOLOv9, and YOLO11 (nano, small, medium, large, and extra-large) were evaluated comparatively. The obtained results are presented in Table 4. When the table is examined, it is seen that the models show a wide range in terms of computational cost and performance metrics. The FLOPs values range from 10.4 to 344.1, the number of parameters ranges from 2.9M to 71.8M, the Precision values range from 79.1% to 90.94%, the Recall values range from 73.13% to 83.74%, and the mAP@50 values range from 82.0% to 86.09%. It clearly shows that there is a non-linear relationship between model scale and performance.

When evaluated in terms of FLOPs, the model with the lowest computational cost is YOLO11n (10.4 GFLOPs), while the highest computational cost is seen in the YOLOv8x model (344.1 GFLOPs). However, it is observed that models with the highest FLOPs values do not provide the highest accuracy performance. It shows that high computational complexity does not yield performance gains when it is not aligned with the scale and diversity of the dataset.

When examined in terms of the number of parameters, the model with the lowest number of parameters is YOLO11n with 2.9M, while the model with the highest number of parameters is YOLOv8x with 71.8M. However, it is clear that the models with the highest number of parameters are not the most successful models in terms of mAP@50. The fact that the YOLOv8s (11.8M) and YOLO11s (10.1M) models, which fall within the 10–12 million parameter range, exhibit more balanced and higher performance.

Table 4. Comparative results of YOLOv8, YOLOv9, and YOLO11 models in terms of computational complexity and detection performance metrics

| Model | Performance measurement metrics | | | | |
|---|---|---|---|---|---|
| | FLOPs | Params(M) | P(%) | R(%) | mAP@50 |
| YOLOv8n | 12.6 | 3.4 | 85.51 | 78.38 | 83.75 |
| YOLOv8s | 42.6 | 11.8 | 88.63 | 78.0 | 86.09 |
| YOLOv8m | 110.2 | 27.3 | 79.1 | 79.79 | 83.39 |
| YOLOv8l | 220.5 | 46.0 | 83.77 | 76.34 | 83.72 |
| YOLOv8x | 344.1 | 71.8 | 84.30 | 78.55 | 82.0 |
| YOLOv9c | 159.4 | 27.9 | 90.94 | 77.68 | 83.56 |
| YOLOv9e | 248.4 | 60.5 | 85.48 | 78.88 | 82.63 |
| YOLO11n | 10.4 | 2.9 | 81.91 | 73.13 | 82.31 |
| YOLO11s | 35.5 | 10.1 | 80.14 | 83.74 | 84.96 |
| YOLO11m | 123.3 | 22.4 | 82.93 | 80.86 | 84.72 |
| YOLO11l | 142.2 | 27.6 | 84.09 | 76.22 | 84.46 |
| YOLO11x | 319.0 | 62.1 | 86.12 | 79.72 | 84.58 |

In terms of the Precision (P) metric, the most successful model was YOLOv9c with 90.94%. In determining the ripeness of strawberries, reducing false positives is crucial because misclassified fruits can lead to errors in harvest timing. In this context, it can be concluded that the YOLOv9c model follows a more conservative and selective prediction strategy. In contrast, the lowest Precision value was observed in the YOLOv8m model at 79.1%. In terms of the Recall (R) metric, YOLO11s was the most successful model with 83.74%. The Recall metric is a key indicator for minimizing the number of misclassified objects, especially in detecting morphologically intermediate classes such as semi-ripe. In this context, it is clear that the YOLO11s model adopts a more sensitive (aggressive) detection strategy. The lowest Recall value was achieved by the YOLO11n model at 73.13%.

The mAP@50 metric is a comprehensive measure representing overall success. The most successful model in terms of this metric was YOLOv8s with 86.09%. The lowest mAP@50 value was achieved by the YOLOv8x model at 82.0%. When performing a general analysis based on model sizes, Nano models (YOLOv8n, YOLO11n) run with low parameters and low FLOPs, but their performance metrics remain limited. Small-scale models (YOLOv8s, YOLO11s) provide the most balanced and highest performance. In medium and large models (v8m, v8l, v9c, v11m, v11l), performance increases reach saturation after a certain point.

Extra-large models (v8x, v9e, v11x), despite their high computational cost, do not offer a meaningful performance advantage.

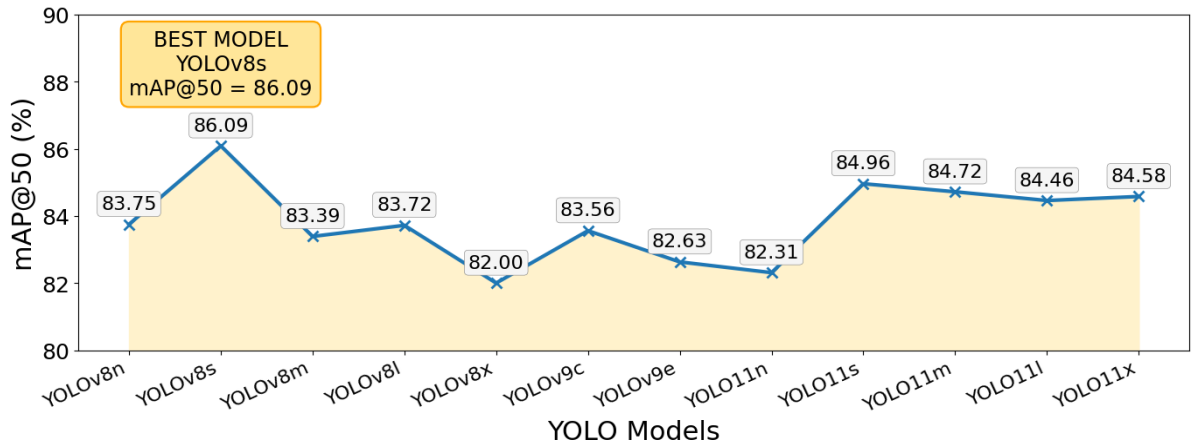

**Figure 6.** Comparison of mAP@50 performances of YOLO models

Figure 6 visually compares the mAP@50 performance of YOLO models. Upon examining the graph, it is clear that there is no linear increase in performance with model scale. The highest mAP@50 value is 86.09%, belonging to the YOLOv8s model, which is also marked as the "best model" on the graph.

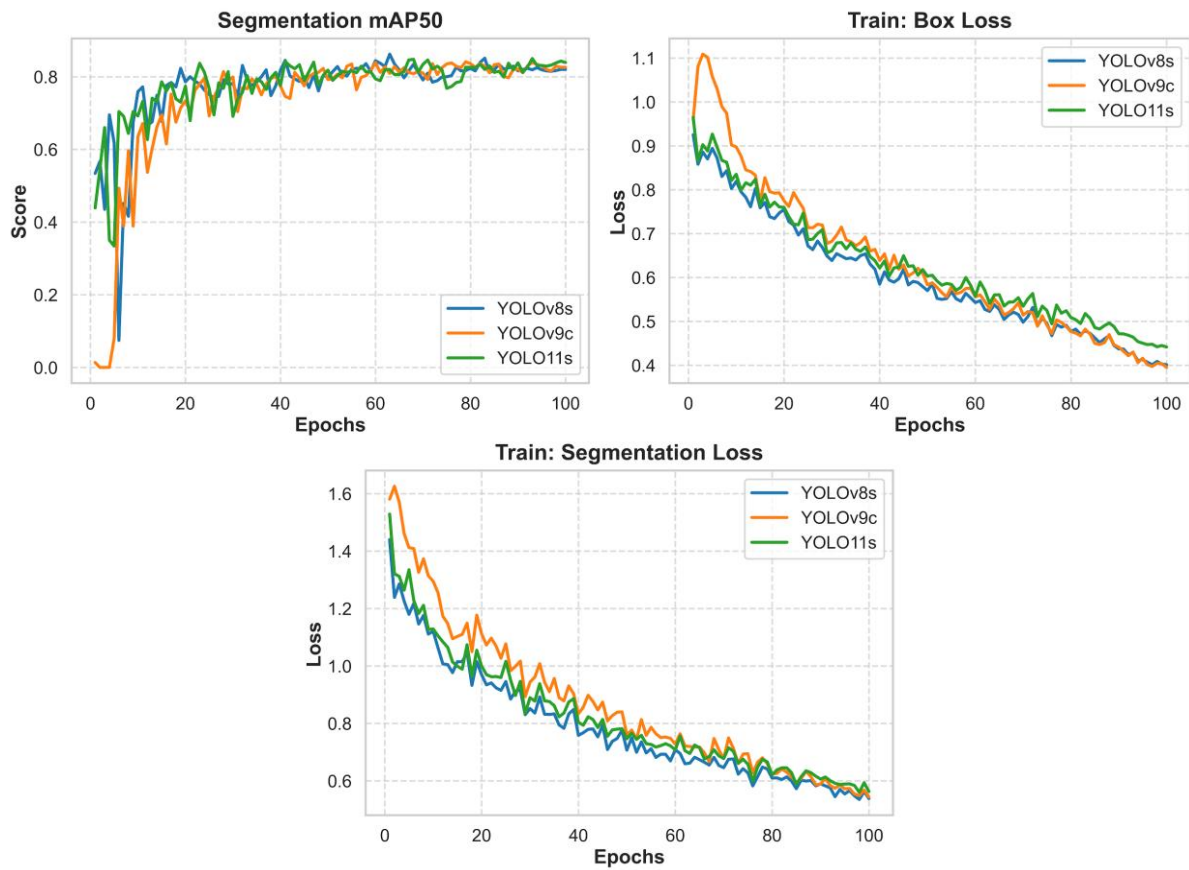

**Figure 7.** Training performance comparison of YOLOv8s, YOLOv9c and YOLO11s models in terms of mAP@50 and loss values

Figure 7 presents the performance curves of the training process for the three most successful models: YOLOv8s, YOLOv9c, and YOLO11s. When examining the graphs, it can be seen that the models achieve similar final accuracy levels but differ in terms of convergence speed and optimization characteristics.

When evaluating the segmentation mAP@50 graph, it is seen that the YOLOv8s and YOLO11s models show a rapid increase within the first 10–15 epochs, reaching the 0.6–0.7 range. The YOLOv9c model, on the other hand, showed low initial performance during the first few epochs but reached a similar level to the other models with a sharp increase after approximately 15–20 epochs. In the later stages of the training process, all three models stabilized in the 0.80–0.83 range and converged to similar final mAP values.

When examining the Box Loss curves, it is observed that the loss values decrease steadily across all models, indicating a stable optimization process. Although YOLOv9c initially produced higher and more fluctuating loss values, it rapidly declined in subsequent epochs to reach levels comparable to other models. YOLOv8s has shown a more balanced and low-variance decrease curve, while YOLO11s has shown a steady decline after a moderate initial loss. At the end of training, the box loss values of the three models ended at very close levels. Segmentation Loss curves also show a similar trend. Although the YOLOv9c model initially had the highest loss value, it showed a significant improvement during the process. YOLOv8s and YOLO11s, on the other hand, exhibited lower initial losses and a more stable decrease curve. In particular, the YOLOv8s model has shown a more stable optimization profile throughout the entire training process.

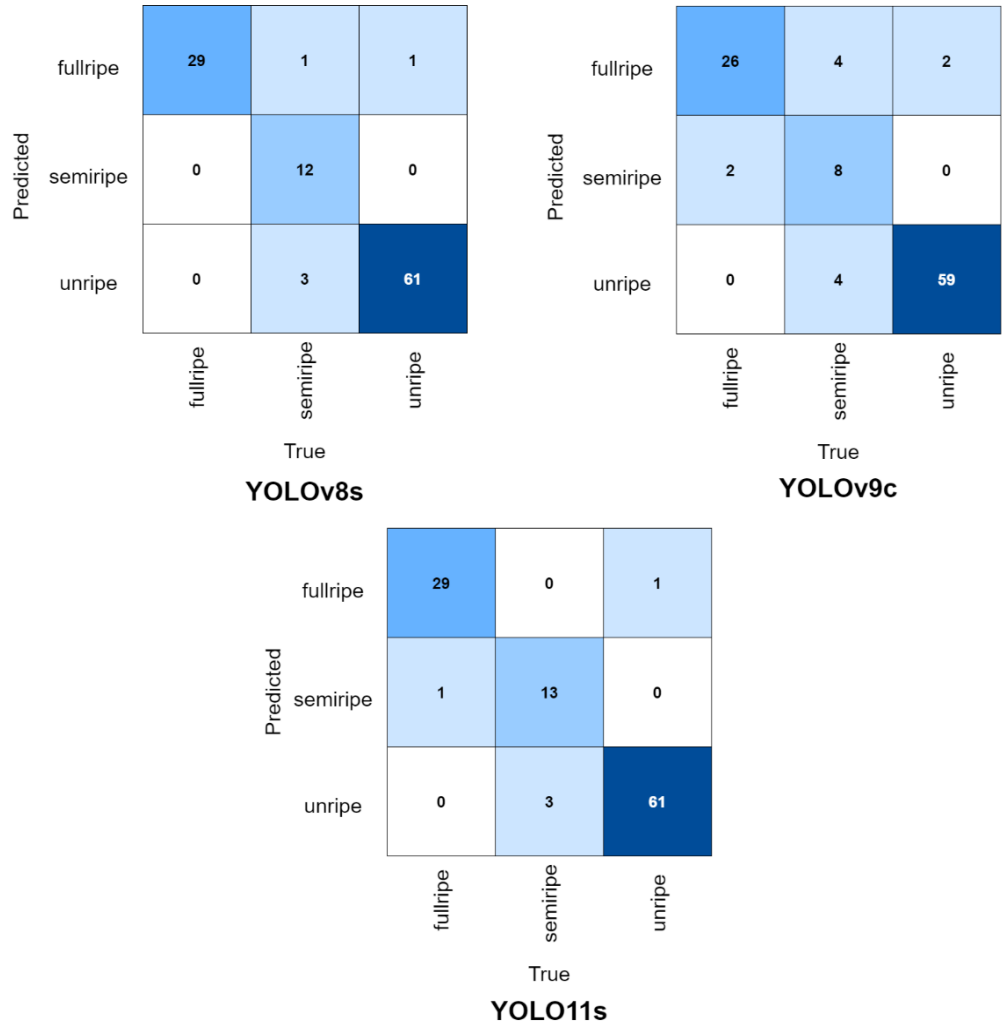

**Figure 8.** Prediction performance analysis of YOLOv8s, YOLOv9c and YOLO11s models

In Figure 8, the class-based prediction performance of the YOLOv8s, YOLOv9c, and YOLO11s models is analysed based on True Positive, False Negative, and False Positive distributions. When the graphs are examined, it is seen that the models diverge in the semi-ripe class.

When evaluated in terms of the fully-ripe class, the YOLO11s and YOLOv8s models achieved the highest True Positive value with 29 correct detections, while YOLOv9c produced 26 correct detections. Furthermore, the YOLO11s model produced only 1 False Positive in this class, indicating that it adopted a stronger prediction approach. YOLOv9c, on the other hand, had the highest false positive rate in this class with 6 False Positives.

The semi-ripe class stands out as the most challenging class due to its morphological transition characteristics. In this class, the YOLO11s model outperformed the other two models with 13 True Positives (compared to 12 for YOLOv8s and 8 for YOLOv9c). Additionally, YOLO11s has a lower False Negative value of 3. In contrast, 4 and 8 False Negatives are observed in the YOLOv8s and YOLOv9c models, respectively. This result shows that the YOLO11s model has higher sensitivity in distinguishing intermediate ripeness levels.

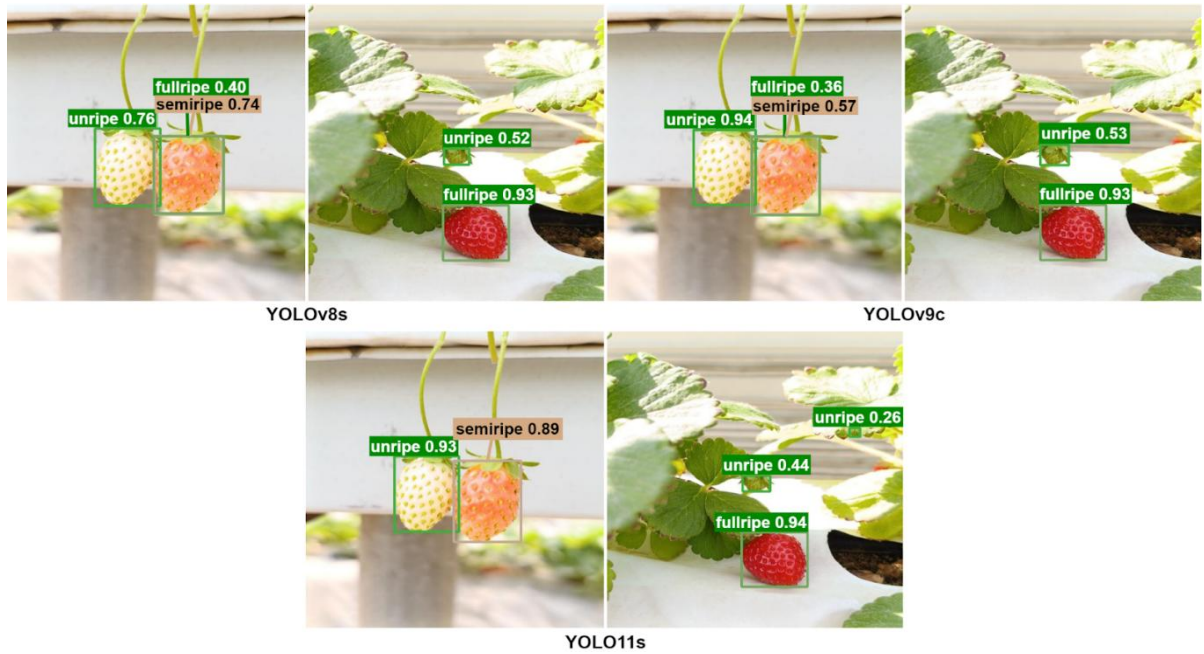

**Figure 9.** Comparison of detection results

In Figure 9, the models' detection results on test images are compared. The YOLOv8s and YOLOv9c models successfully detected clearly ripe and unripe strawberries but produced low confidence scores on semi-ripe samples and occasionally misclassified them. The YOLO11s model correctly distinguished the semi-ripe class with a higher confidence value but produced low-confidence false positives in some background regions.

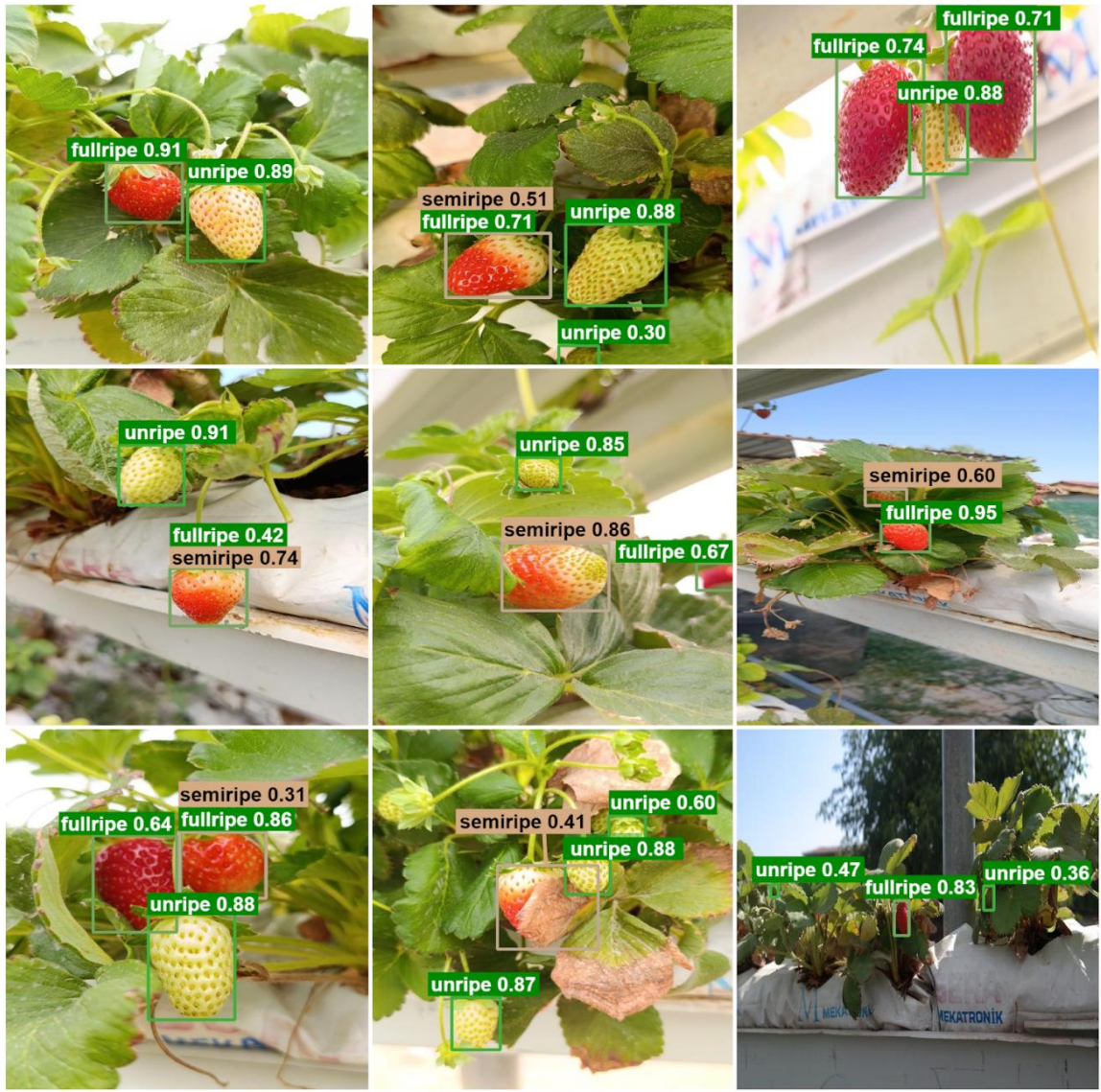

**Figure 10.** YOLOv8s predictions

In Figure 10, the prediction results of the YOLOv8s model on different test images are presented. When the images are examined, it is seen that the model makes successful detections with high confidence scores in the fully-ripe and unripe classes. The classification accuracy is particularly high in examples with distinct color differences. In contrast, it is remarkable that the confidence scores are relatively low in the semi-ripe class and show a tendency to mix with fully-ripe in some examples. Furthermore, low-confidence predictions and occasional class transition errors occur in scenes with leaf overlap, shadows, and complex backgrounds. Overall, the YOLOv8s model demonstrates high accuracy in situations with clear visual distinctions; however, it may struggle to distinguish between classes in transitional ripeness levels and complex environmental conditions.

## 5. Discussion

In this study, the YOLOv8, YOLOv9, and YOLO11 architectures were systematically compared on a new, publicly available strawberry ripeness dataset collected from two different greenhouses under varying light conditions. The findings clearly demonstrate a non-linear relationship between model scale and performance. Specifically, small and medium-scale models in the 10–12 million parameter range were found to yield the most balanced results in terms of accuracy and computational cost. One of the most notable results of the study is that the highest mAP@50 value (86.09%) was achieved by the YOLOv8s model. This finding shows that smaller but balanced architectures can generalize better on relatively limited datasets with class imbalance. The highest precision result (90.94%) was achieved with the YOLOv9c model, while the highest recall value (83.74%) was provided by the YOLO11s model. This situation shows that architectural design preferences (e.g., more conservative or more aggressive prediction strategies) are decisive on the precision–recall balance. In particular, the higher recall value of YOLO11s reveals that it offers a more sensitive structure in the detection of morphologically intermediate classes, such as semi-ripe ones. In qualitative analyses, it was also observed that YOLO11s made more accurate classifications in samples with subtle color transitions. Conversely, the same model's tendency to produce more false positives in complex backgrounds suggests that high sensitivity may sometimes come at the expense of precision. Many studies in the literature report performance improvements by optimizing a single architecture; however, systematic comparisons of different generations of YOLO models under the same dataset and experimental conditions are limited. The results obtained in this study show that the assumption that "a larger model provides better performance" is not valid for every dataset. In particular, the fact that higher-parameter models such as YOLOv8x and YOLOv9e produce lower mAP values compared to smaller variants highlights the need for model complexity to be aligned with the scale of the dataset.

Furthermore, the frequent use of private datasets in the literature makes direct comparisons between studies difficult. In this context, presenting a publicly available dataset collected from two different greenhouse environments is an important contribution in terms of methodological transparency and reproducibility. The reflection of real-world challenges such as varying light conditions, leaf overlap, and fruit density in the dataset enhances the transferability of the results to practical applications.

One of the unexpected findings in the study is that performance declines beyond a certain point as the number of parameters increases. In particular, the fact that large models such as YOLOv8x and YOLO11x achieve lower mAP scores than smaller variants suggests that the size and diversity of the dataset does not provide sufficient representativeness for high-capacity models. This can be explained by an increase in overfitting tendency or an increase in the model's sensitivity to noise. Another important observation is that YOLOv9c does not rank first in mAP@50 despite its high precision. This shows that precision alone does not represent overall performance; when balanced with recall, average success metrics may be limited. Therefore, it is understood that model selection should be made depending on the application scenario (for example, minimizing false negatives may be a priority in harvesting robots). This study has some limitations. First, although the dataset was collected from two different greenhouses, the geographical and climatic diversity is limited. Greenhouse types in different countries, open field production conditions, or different strawberry varieties may present different distributions to the model. Second, there is a significant imbalance between classes (the unripe class is dominant). Although this reflects real production conditions, it may have made it difficult to learn decision boundaries, especially in the semi-ripe class. In future studies, expanding the dataset with images collected from different geographical regions and different production systems (open field, vertical farming, different cultivars) will increase the model's generalizability. Additionally, data augmentation strategies and class-weighted loss functions can be tried to reduce the class imbalance problem.

## 6. Conclusion

This study has introduced a new publicly available strawberry ripeness dataset to the literature, created using images collected from two different greenhouse environments and varying light conditions, and comprehensively compared the YOLOv8, YOLOv9, and YOLO11 architectures under the same experimental conditions. The results show that there is no linear increase in performance with model scale; instead, compact models in the 10–12 million parameter range (especially YOLOv8s and YOLO11s) offer the most suitable performance in terms of accuracy–sensitivity–computational cost balance. This finding indicates that architectural efficiency should be prioritized over high hardware requirements in smart farming applications and real-time harvesting systems. One of the most important contributions of the study is that it provides a repeatable and comparable experimental framework. However, factors such as the limited geographical scope of the dataset and the imbalance in class distribution should be considered in terms of the generalizability of the results.

Future studies conducted with different production systems, different varieties, and larger-scale datasets will contribute to the validation of the proposed findings at the field level. In conclusion, this research not only presents a model comparison but also establishes a practical and methodological reference framework for the architectural scale-performance relationship in strawberry ripeness detection. In this respect, it constitutes a fundamental step towards the development of real-time and reliable ripeness detection systems.

**Declarations**

**Author Contributions**

All authors contributed equally to this work. All authors have read and agreed to the published version of the manuscript.

**Conflict of Interest**

The authors declare no potential conflicts of interest with respect to the research, authorship, and/or publication of this article.

**Data availability**

The dataset can be download from following link.

https://www.kaggle.com/datasets/mahyeks/multi-class-strawberry-ripeness-detection-dataset

**Funding**

This research did not receive any specific grant from funding agencies in the public, commercial, or not-for-profit sectors.